\documentclass[letterpaper, 10 pt, journal, twoside]{IEEEtran}  

\IEEEoverridecommandlockouts                              

\usepackage{cite}
\usepackage{comment}
\usepackage[nolist]{acronym}
\usepackage{amsmath}
\usepackage{amssymb}
\usepackage{bm}
\usepackage{amssymb}
\usepackage{booktabs}
\usepackage{rotating}
\usepackage[detect-weight=true]{siunitx}  
\usepackage[final]{microtype}
\usepackage[font=small,labelfont=bf]{caption}
\usepackage{subcaption}
\usepackage{tablefootnote}
\usepackage{multirow}
\usepackage[hidelinks]{hyperref}
\usepackage[normalem]{ulem}
\usepackage{xcolor}

\newcommand{\maplab}{\texttt{maplab~2.0} }
\newcommand{\maplabns}{\texttt{maplab~2.0}}

\newcommand{\rotHeader}[1]{\begin{sideways}\textbf{#1}\end{sideways}}
\newcommand{\m}{\,\si{\metre}}
\newcommand{\km}{\,\si{\kilo\metre}}

\newcommand{\iconmono}{\includegraphics[height=1.2em]{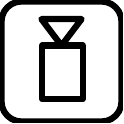}}
\newcommand{\iconstereo}{\includegraphics[height=1.2em]{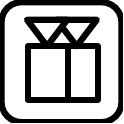}}
\newcommand{\iconimu}{\includegraphics[height=1.2em]{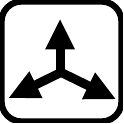}}
\newcommand{\iconlidar}{\includegraphics[height=1.2em]{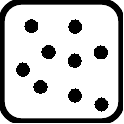}}

\newcommand{\iconmonotext}{\includegraphics[height=1em]{imgs/maplab_icon_mono.pdf}}
\newcommand{\iconstereotext}{\includegraphics[height=1em]{imgs/maplab_icon_stereo.pdf}}
\newcommand{\iconimutext}{\includegraphics[height=1em]{imgs/maplab_icon_imu.pdf}}
\newcommand{\iconlidartext}{\includegraphics[height=1em]{imgs/maplab_icon_lidar.pdf}}

\newcommand{\reviewchanges}[1]{{#1}} 
\renewcommand{\sout}[1]{}
\newcommand{\oldeq}[1]{}

\title{\LARGE \bf
\maplab -- A Modular and Multi-Modal Mapping Framework
}

\author{Andrei Cramariuc$^{1,\ast}$, Lukas Bernreiter$^{1,\ast}$, Florian Tschopp$^{2,\ast}$, Marius Fehr$^{3}$, Victor Reijgwart$^{1}$\\ Juan Nieto$^{4}$, Roland Siegwart$^{1}$, Cesar Cadena$^{1}$
\thanks{\reviewchanges{This work was supported in part by the National Center of Competence in Research (NCCR) on Digital Fabrication and received Funding from the European Union’s Horizon 2020 Research and Innovation Programme Under Grant Agreement 871542.}}%
\thanks{$^{\ast}$ Authors contributed equally to this work.}
\thanks{$^{1}$Authors are members of the Autonomous Systems Lab, ETH Zurich, Switzerland; {\tt\small \{firstname.lastname\}@mavt.ethz.ch}}%
\thanks{$^{2}$Author is with Arrival Ltd, UK but the work was done while the author was a member of $^1$; {\tt\small tschopp@arrival.com }}%
\thanks{$^{3}$Author is with Voliro, Switzerland but the work was done while the author was a member of $^1$; {\tt\small marius.fehr@voliro.com}}%
\thanks{$^4$Author is with Microsoft, Switzerland but the work was done while the author was a member of $^1$; {\tt\small juannieto@microsoft.com}}%
\thanks{Special thanks to Thomas Schneider, Marcin Dymczyk, Juichung Kuo, Patrick Pfreundschuh, Nicolas Scheidt, Marius Brühlmeier, and Benjamin Hahn for their significant contributions to the conceptualization and implementation of the framework.}%
}

\begin{document}

\maketitle

\begin{abstract}
Integration of multiple sensor modalities and deep learning into \ac{slam} systems are areas of significant interest in current research.  
Multi-modality is a stepping stone towards achieving robustness in challenging environments and interoperability of heterogeneous multi-robot systems with varying sensor setups.
With \maplabns, we provide a versatile open-source platform that facilitates developing, testing, and integrating new modules and features into a fully-fledged \ac{slam} system.
Through extensive experiments, we show that \maplabns's accuracy is comparable to the state-of-the-art on the HILTI 2021 benchmark.
Additionally, we showcase the flexibility of our system with three use cases: i) large-scale ($\sim$10\,$\mathrm{km}$) multi-robot multi-session (23 missions) mapping, ii) integration of non-visual landmarks, and iii) incorporating a semantic object-based loop closure module into the mapping framework.
The code is available open-source at \url{https://github.com/ethz-asl/maplab}.
\end{abstract}

\section{Introduction}


\IEEEPARstart{S}{imultaneous} Localization And Mapping (SLAM) is an essential component for various robotic applications, such as autonomous driving~\cite{bresson2017simultaneous}, mobile manipulation~\cite{Blomqvist2020GoEnvironments}, and augmented/mixed reality.
In these applications, the robotic platform needs to be aware of the surrounding environment and its location to perform the given task, be it driving autonomously to a particular destination or picking up and delivering an object.
One step further is the ability to perform long-term mapping, which typically requires tools for processing and merging multiple maps enabling an even wider range of diverse applications and tasks.


Over recent years, many tailored \ac{slam} solutions have been successfully developed for specific environments or sensor configurations~\cite{ labbe2019rtab, Rosinol2020Kimera:Mapping, Tian2022, chang2022lamp, karrer2018cvi, campos2021orb, schmuck2021covins, lajoie2020door, Schneider2017}.
However, many challenges remain until \ac{slam} is fully solved or generically deployed in ubiquitous operating conditions.
Recent efforts in fusing multiple modalities have gained significant traction due to the ability of multi-modal systems to compensate for weaknesses in individual sensors or methods.
Thus, enabling a more robust robotic operation in degraded environments and even with full or partial sensor failures.
Works combining many different sensors exist~\cite{labbe2019rtab, Rosinol2020Kimera:Mapping, Tian2022} and achieve remarkable performance.
However, together with other open-source \ac{slam} frameworks~\cite{campos2021orb, lajoie2020door, schmuck2021covins, Schneider2017}, these systems are tightly integrated.
More specifically, they function only with specific sensor configurations, and the basic modules (\textit{e.g.}, odometry, localization, or feature extraction) are highly entangled.
Modifying those modules or incorporating new functionalities requires significant engineering work, adding major overheads to scientific research and the development of new products.
\reviewchanges{\sout{In these cases}Therefore}, versatile systems that can seamlessly integrate various sensor setups and leverage multiple sensor modalities are desirable.
Flexible support of multiple modalities is also the stepping stone for heterogeneous multi-robot systems, where different robots \reviewchanges{\sout{might}can} be equipped with varying combinations of sensors, \textit{e.g.}, due to platform constraints.

\begin{figure}[t]
\centering
	\includegraphics[width=1.0\columnwidth, trim= 0mm 0mm 0mm 0mm, clip]{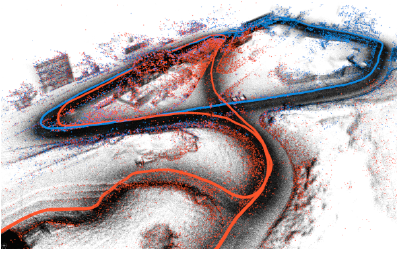}
	\caption{We propose \maplabns, a flexible and generic multi-robot, and multi-modal framework. \maplab can seamlessly integrate multiple robots (colored paths), visual landmarks (colored points), and LiDAR scans (black points).}
	\label{fig:intro:teaser} 
	\vspace{-0.7cm}
\end{figure}


\maplab provides an open-source platform for multi-session, multi-robot, and versatile multi-modal mapping.
The original \textit{maplab}~\cite{Schneider2017} was an open-source toolbox for creating and managing exclusively visual-inertial maps.
With \maplabns, we extend the original framework far beyond its initial scope by integrating multiple new modalities such as LiDAR, GPS receivers, wheel encoders, semantic objects, and more.
These examples provide the templates for easy extension to further sensing modalities.
\maplab also offers interfaces for easy integration of external components, such as adding any number of different visual features or loop closure constraints.
These features make our new platform ideally suited as a development and research tool for deep-learned keypoint detectors and loop closure engines that, until now, have mostly been tested separately~\cite{Detone2018SuperPoint:Description}.
Additionally, online collaborative SLAM is now possible in \maplab due to the new submapping capabilities, enabling online building, optimization, and co-localization of one global map from multiple sources.
This is made possible by our implementation of a new centralized server node that aggregates the data from multiple robots and can transmit the collaboratively built map back to the robots for increased performance~\cite{bernreiter2022collaborative}.
We showcase the capabilities and performance of our system in multiple experiments and datasets, providing proof of concept implementations for non-visual keypoints, deep-learned descriptor integration, and a semantic object-based loop closure engine.

Our contributions can be summarized as follows:
\begin{itemize}
    \item We provide an open-source, multi-modal, and multi-robot mapping framework that allows integration and fusion of an unparalleled amount of different data compared to other existing methods.
    \item An online collaborative mapping system that utilizes submapping and a central server to compile and distribute globally consistent, feature-rich maps.
    \item Integration of interfaces for any number of custom feature points, descriptors, and loop closure. We showcase their flexibility in experiments featuring  3D LiDAR keypoints and semantic object-based loop closures.
\end{itemize}

\begin{figure*}[!t]
\centering
	\includegraphics[width=1.0\textwidth, trim= 0mm 0mm 0mm 0mm, clip]{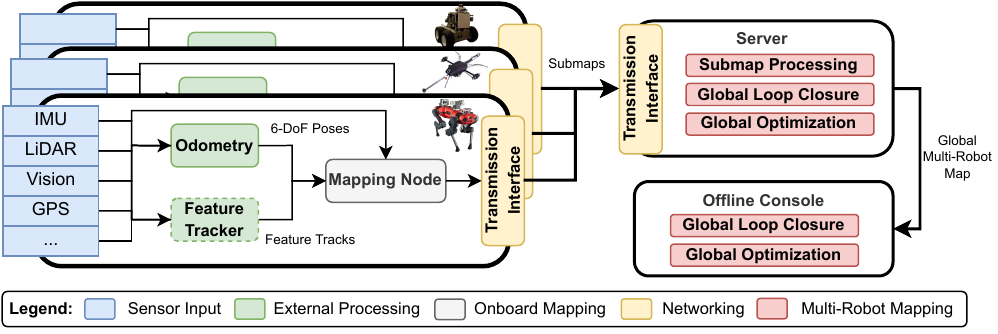}
	\caption{Overview of the \maplab framework and its three main components, namely the mapping node, the centralized server, and the offline console. The mapping node runs on each robot and collects the sensor data into submaps which are passed to the centralized server that merges them into a globally consistent map. This map can then, at later stages, be refined or merged with other maps using the tools provided by the offline console. This figure showcases an example of one possible configuration of how \maplab can be used. There are multiple other combinations as well as inputs and modules that can be added or excluded as described in Section~\ref{sec:maplab}.}
	\label{fig:maplab:overview}
	\vspace{-0.3cm}
\end{figure*}

\section{Related Work}
Mapping can be defined as the challenge of creating environment representations and has seen a vast and diverse range of solutions over the past decades, with significant changes driven by new sensors and scenarios~\cite{cadena2016past}.
Multi-modality has evolved beyond standard sensor fusion (\textit{i.e.}, visual-inertial or stereo cameras) to include more complex combinations involving, for example, LiDARs and semantic information.
Another notable topic is multi-robot mapping, where multiple robots simultaneously explore an environment and aim to create one globally consistent map.
Multi-robot mapping differs from multi-session mapping, which involves collecting measurements of the same place at separate time intervals and enabling offline operations to and between sessions.
While multi-robot frameworks can be used in a multi-session manner by sequentially processing data recordings, this is inefficient as it requires reprocessing all previous data any time a new recording is added due to their lack of map management tools.
A comparison of significant \ac{slam} frameworks and their features is presented in Table~\ref{tab:related_work:sota_feature_comparison}.

The first version of \textit{maplab}~\cite{Schneider2017} is a multi-session mapping framework designed for visual-inertial systems.
Other comparable frameworks are ORB-SLAM3~\cite{campos2021orb} and RTAB-Map~\cite{labbe2019rtab}.
ORB-SLAM3 is an extension of its predecessor ORB-SLAM2~\cite{Mur-Artal2017}, adding support for an IMU and multi-session mapping capabilities.
RTAB-Map integrates vision and depth measurements from a LiDAR or an RGB-D camera.
An extension~\cite{labbe2022multi} to RTAB-Map supports a variety of handcrafted visual features and SuperPoint~\cite{Detone2018SuperPoint:Description}, but does not allow for easy integration of other descriptors.
Both frameworks offer similar map creation and management features as \textit{maplab}, with the addition of online loop closure and optimization during mapping.
All three of the aforementioned frameworks are tightly integrated systems designed for a particular sensor configuration.
On the contrary, we allow easy integration of different sensor setups, visual features, and support an arbitrary odometry input in \maplabns, which \reviewchanges{\sout{not only }}facilitates the use of heterogeneous robots \reviewchanges{\sout{but also}and} provides a new level of flexibility.

Kimera~\cite{Rosinol2020Kimera:Mapping} is a multi-modal mapping framework that provides both local and global 3D meshes with semantic annotations and a global trajectory estimate based on visual-inertial \ac{slam}.
As opposed to \maplabns, Kimera does not have multi-session capabilities, and the 3D reconstruction with its semantic annotations is not used to improve the accuracy of the \ac{slam} estimates.
In general, semantic information has the potential to significantly improve mapping~\cite{cadena2016past} by being a catalyst for high-level scene understanding.
However, previous works utilizing semantics for mapping~\cite{dube2020segmap,Gawel2017,Cramariuc2021SemSegMapLocalization,Taubner2020LCDRecognition,bernreiter2019multiple} focus mainly on generating improved descriptors rather than leveraging them in a fully semantic \ac{slam} system.
In this work, we propose utilizing image descriptors and a simple semantic object representation, which allows us to optimize using well-known relative pose errors.

\begin{table}[!t]
    \vspace{0.18cm}
    \setlength{\tabcolsep}{3pt}
    \centering
    \caption{Comparison of supported features in state-of-the-art mapping frameworks. (\textit{Diff. sensors:} Maps with different sensor configurations can be combined; \textit{Ext.}: External source; \textit{LC.}: Loop closure; $\mathit{\ast}$: IMU biases have to be provided alongside the poses)}
    \label{tab:related_work:sota_feature_comparison}
    \begin{tabular}{c|ccccccccccc}
        \toprule
        & \rotHeader{Multi-Modal} & \rotHeader{Multi-Robot} & \rotHeader{Multi-Session} & \rotHeader{Online} & \rotHeader{Diff. Sensors} & \rotHeader{Ext. Odometry} & \rotHeader{Ext. Features} & \rotHeader{Ext. LCs} & \rotHeader{GPS support} & \rotHeader{Semantic LCs} & \rotHeader{Open-Source}\\
        \midrule
        RTAB-Map~\cite{labbe2019rtab} & \checkmark & \quad & \checkmark & \checkmark & \quad & \checkmark & \quad & \quad & \quad & \quad & \checkmark \\
        ORB-SLAM3~\cite{campos2021orb} & \checkmark & \quad & \checkmark & \checkmark & \quad & \quad & \quad & \quad & \quad & \quad & \checkmark \\ 
        LAMP 2.0~\cite{chang2022lamp} & \checkmark & \checkmark & \quad & \checkmark & \checkmark & \checkmark & \quad & \quad & \quad & \quad & \checkmark \\ 
        CVI-SLAM~\cite{karrer2018cvi} & \quad & \checkmark & \quad & \checkmark & \quad & \quad & \quad & \quad & \quad & \quad & \quad \\ 
        COVINS~\cite{schmuck2021covins} & \quad & \checkmark & \quad & \checkmark & \quad & $\ast$ & \quad & \quad & \quad & \quad & \checkmark \\ 
        DOOR-SLAM~\cite{lajoie2020door} & \checkmark & \checkmark & \quad & \checkmark & \quad & \checkmark & \quad & \quad & \quad & \quad & \checkmark \\ 
        Kimera~\cite{Rosinol2020Kimera:Mapping} & \checkmark & \quad & \quad & \checkmark & \quad & $\ast$ & \quad & \quad & \quad & \quad & \checkmark \\ 
        Kimera-Multi~\cite{Tian2022} & \checkmark & \checkmark & \quad & \checkmark & \quad & $\ast$ & \quad & \quad & \quad & \quad & \checkmark \\ 
        maplab~\cite{Schneider2017} & \quad & \quad & \checkmark & \quad & \quad & \quad & \quad & \quad & \quad & \quad & \checkmark \\ 
        \maplab & \checkmark & \checkmark & \checkmark & \checkmark & \checkmark & \checkmark & \checkmark & \checkmark & \checkmark & \checkmark & \checkmark \\ 
        \bottomrule
    \end{tabular}
    \vspace{-0.6cm}
\end{table}

Kimera-Multi~\cite{Tian2022} is a direct extension of Kimera~\cite{Rosinol2020Kimera:Mapping}, which enables the multi-robot scenario using a fully distributed system but does not improve on the purely visual-inertial \ac{slam} backend in the original Kimera.
Our approach instead uses a centralized server that collects submaps, optimizes them, and creates one globally consistent map.
A similarly centralized setting was also explored by COVINS~\cite{schmuck2021covins}.
However, COVINS is limited to the visual-inertial use case, while \maplab can incorporate multiple sensor modalities and configurations\reviewchanges{\sout{ into the same global map}}.
In a similar vein are also LAMP 2.0~\cite{chang2022lamp}, CVI-SLAM~\cite{karrer2018cvi}, and DOOR-SLAM~\cite{lajoie2020door}, which offer collaborative mapping between robots but are tightly integrated systems, limited to one sensor modality, and have little flexibility.

Although various other \ac{slam} frameworks exist, they are mainly focused on specific sensor or robot-environment configurations, and changes to either one are usually difficult or impossible.
To our knowledge of all the existing methods, \maplab is the most flexible mapping and localization framework that not only supports a variety of sensors but can also be seamlessly adapted to new needs.

\section{The \maplab Framework} \label{sec:maplab}
The general structure of the \maplab framework is presented in Figure~\ref{fig:maplab:overview}.
The entire framework can be divided into three main components: the mapping node, the mapping server, and the offline console interface.
We begin with an overview of the underlying map structure in \maplabns, after which we discuss in more detail the main components.

\subsection{Map Structure}
\label{sec:map-structure}
We denote a \textit{map} as a collection of one or more \textit{missions}, where each mission is based on a single continuous mapping session.
The underlying structure of a map is a factor graph consisting of vertices and edges that incorporate all the robot information and the measurements across different missions.
The state of the robot at a certain point in time $t$ is parameterized as a vertex (6~\ac{dof} pose, velocity, IMU biases).
Landmarks are also represented as vertices in the graph whose state is defined as a 3D position.
The 3D landmarks can be used as an underlying representation for anything in the environment with a 3D position, \textit{e.g.}, visual landmarks, 3D landmarks, or even semantic objects.

\subsubsection{Constraints} Vertices are connected through different types of edges that impose constraints on their state variables based on observations (\textit{e.g.}, keypoints, imu measurements, and loop closures).
IMU edges contain the pre-integrated IMU measurements between connected vertices and therefore only connect temporally sequential vertices.
Relative pose constraint edges impose a rigid 6~\ac{dof} transformation between two vertices and are used to represent either relative motion (\textit{i.e.}, odometry) or loop closures across larger temporal gaps or missions.
The edges are assigned a covariance to quantify the measurement noise, which is typically set to a predefined constant value.
The covariance can be used to model the degrees of motion a sensor can observe, \textit{e.g.} wheel odometry has infinite covariance for motion along the z-axis, as well as pitch and roll.
We consider loop closure edges a special case of relative pose constraint edges.
For increased robustness and to account for outliers, loop closure edges can be included as switchable constraints~\cite{sunderhauf2012switchable}.
The optimizer can then discard \reviewchanges{\sout{individual }}edges from the graph if they conflict too much with the other constraints.
Finally, edges connect a landmark to the poses from which it was observed and impose an error based on the difference between the estimated and observed landmark position.

During optimization, constraints can also be imposed directly on the internal states of chosen vertices.
For example, absolute constraints enforce a global 3D position on a vertex with a given uncertainty and allow us to integrate GPS measurements or absolute fiducial marker observations.
Additionally, fixing certain states enables the flexibility of choosing which parts of the problem are to be optimized.


\subsubsection{Landmarks}\label{sec:landmarks} The visual mapping module at the core of \textit{maplab}~\cite{Schneider2017} is still a part of \maplabns.
It includes feature detection based on ORB~\cite{Rublee}, with binary descriptors from either BRISK~\cite{Leutenegger2011} or FREAK~\cite{alahi2012freak}.
Feature correspondences between consecutive frames are established based on descriptor matches, where for robustness, the matching window is restricted by integrated gyroscope measurements.
These feature tracks are then triangulated into 3D landmarks.

Global localization and loop closure is done by taking individual frames and establishing a set of 2D-3D matches using the feature descriptors.
A covisibility check is applied afterward to the matches to filter outliers.
Then, with a P3P algorithm within a RANSAC scheme, the remaining matches are used to obtain a transformation with respect to the map's reference frame.
This transformation can then be added to the factor graph as a loop closure edge.
We also provide an alternative method that incorporates loop closures by merging the covisible landmarks and minimizing their reprojection error.
This approach foregoes the difficulty of tuning explicit loop closure edge covariances but enforces softer constraints on the factor graph.

In \maplabns, we have added the possibility of concurrently including any number of different types of features into the map.
To obtain feature tracks across consecutive frames, users can either use the included generic implementation of a Lucas–Kanade tracker~\cite{Lucas1981} or supply the track information themselves.
In addition, we expanded the matching engine to support floating point descriptors, enabling loop closure using the latest developed descriptors.
Binary descriptors are matched, as in \textit{maplab}, using an inverted multi-index~\cite{Lynen2015GetLocalization}, while floating point descriptors are matched using a \ac{flann}~\cite{muja2009fast}.
Matches are then treated similarly for the purpose of loop closures as previously described.
Still, for tuning purposes, different feature types can have separate parameter sets to account for differences in quality and behavior.

\maplab can also handle landmarks with 3D observations.
These could originate from, for example, RGB-D cameras, where visual features also have an associated depth, or from features detected directly in a 3D point cloud.
The significant difference is that the position of these landmarks is not triangulated using multi-view geometry but by averaging the 3D measurements.
Similarly, the pose graph error term is not based on the reprojection error, but on the Euclidean distance between the observed 3D position and the 3D position of the landmark.
The other significant difference is that loop closure is set up as a 3D to 3D RANSAC matching problem without the P3P algorithm.

\subsection{Mapping Node}
The mapping node runs onboard each robot and uses external input sources and the raw sensor data to create a map in the form of a multi-modal factor graph.
A 6~\ac{dof} odometry input is used during map building to initialize the robot pose vertices for the underlying factor graph.
The mapping node is agnostic to the odometry method and features a simple interface, thus enabling its easy use across various robots and sensor setups.
This is in contrast to \textit{maplab}~\cite{Schneider2017}, where only the built-in visual-inertial estimator ROVIOLI~\cite{bloesch2017iterated} was available -- whereas \maplab does not even require an IMU.
However, if an IMU is available, inertial constraints are added to the map, and the state estimator can optionally also compute an initial estimate for the IMU biases.
These bias estimates can then be used to improve the initialization of the global map optimization problem, which benefits its convergence speed and accuracy.
If an IMU is present but not used by the state estimator, the bias estimation can also be done separately~\cite{Lynen2013, Madgwick2011, Valenti2015}.

Substantial changes to the original \textit{maplab} framework were also made such that other sensor modalities can be processed and integrated using custom internal components or easily configurable external interfaces.
Most notably, \maplab can incorporate any number of different 3D landmarks types at runtime.
Furthermore, relative constraints (\textit{e.g.}, odometry or external loop closures) and absolute 6~\ac{dof} constraints (\textit{e.g.}, GPS or fiducial markers) can now seamlessly be added.

The raw camera images or the LiDAR point clouds can be attached to the map as resources that later modules can use at any time, for example, to compute additional loop-closures or detect objects.
The resulting maps with all the included constraints can then be passed on to the mapping server for online processing or stored and loaded for later offline processing in the console.

\begin{figure*}[!htb]
\centering
     \begin{subfigure}[b]{0.49\linewidth}
         \centering
         \includegraphics[width=\textwidth]{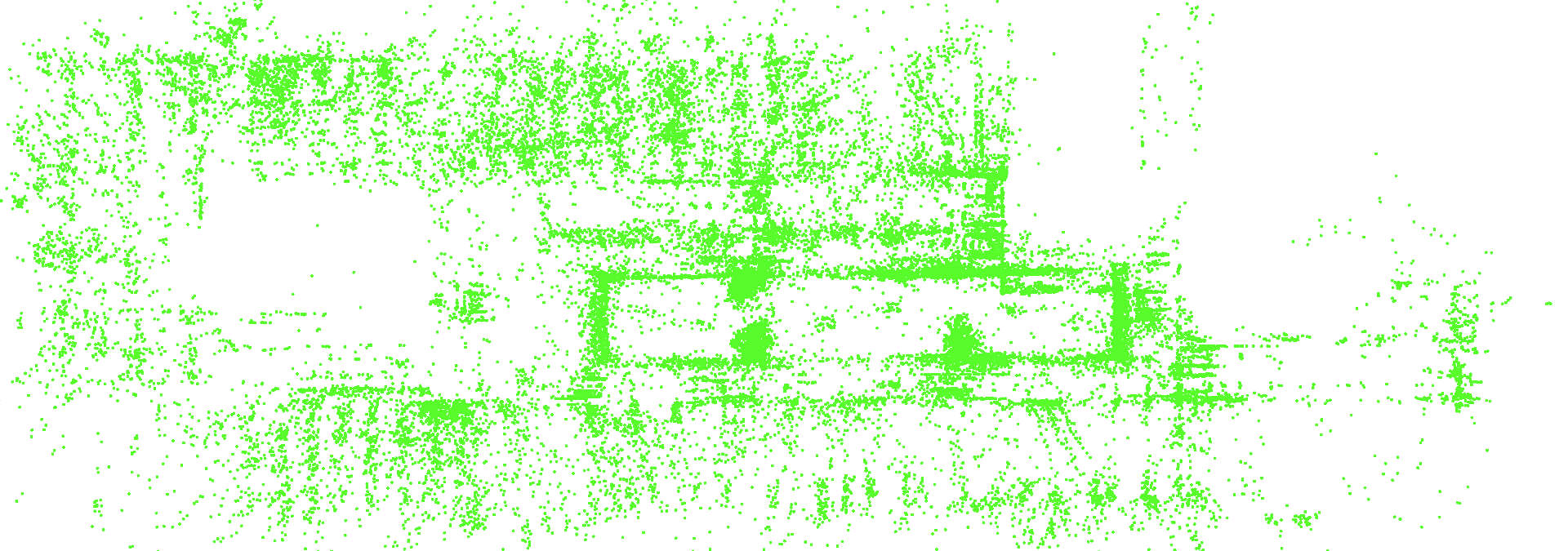}
         \caption{BRISK descriptors with ORB detector.}
     \end{subfigure}
     \hfill
     \begin{subfigure}[b]{0.49\linewidth}
         \centering
         \includegraphics[width=\textwidth]{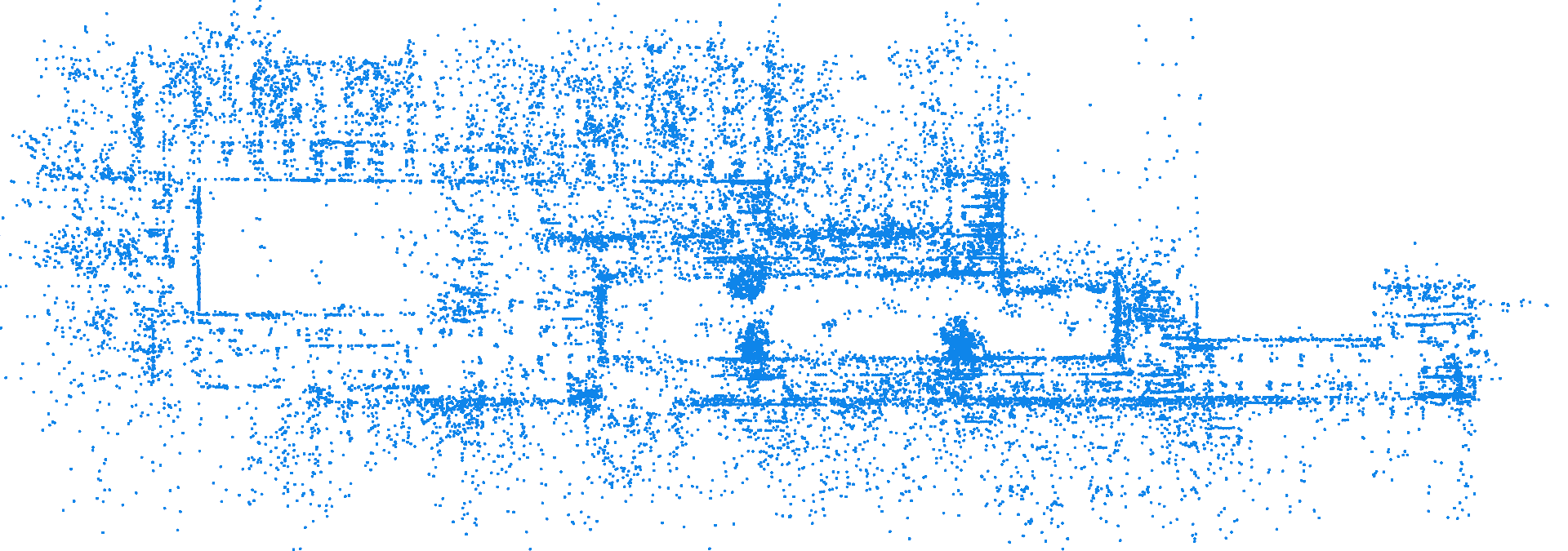}
         \caption{SuperPoint visual features with SuperGlue tracking.}
     \end{subfigure}
     \begin{subfigure}[b]{0.49\linewidth}
         \centering
         \includegraphics[width=\textwidth]{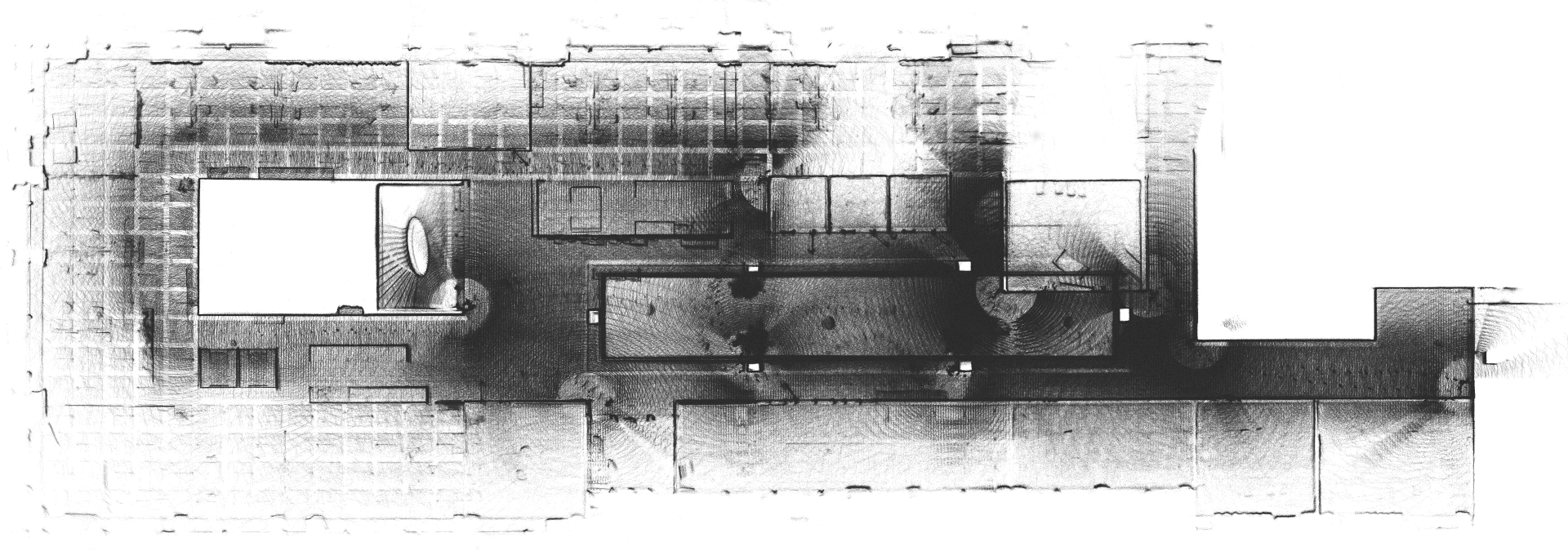}
         \caption{Accumulated point cloud reconstruction.}
     \end{subfigure}
     \hfill
     \begin{subfigure}[b]{0.49\linewidth}
         \centering
         \includegraphics[width=\textwidth]{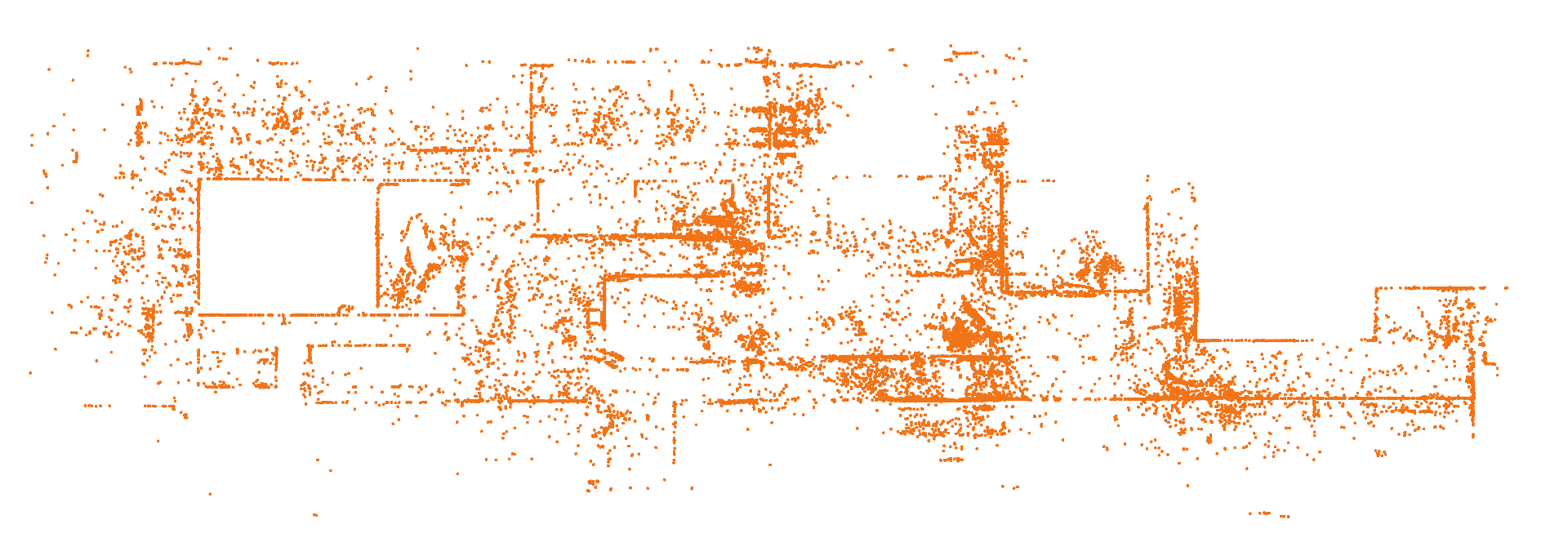}
         \caption{SuperPoint on LiDAR images with SuperGlue tracking.}
         \label{fig:hilti-lidar-kps}
     \end{subfigure}
     \caption{Visualizations of the features and sensor data in \maplab on the Office Mitte sequence from the HILTI 2021 dataset, using OKVIS and global bundle adjustment over the features.}
     \label{fig:hilti-office-mitte}
     \vspace{-0.6cm}
\end{figure*}

\subsection{Mapping Server}
The mapping server is a new addition to \maplabns, enabling collaborative and online mapping.
This method was successfully deployed in the DARPA Subterranean Challenge as part of the winning team's (CERBERUS) multi-robot mapping system~\cite{tranzatto2022science}.
The server node can run on a dedicated machine or one of the robots in parallel with the mapping node.
The mapping nodes divide their maps into chunks, called submaps, at regular intervals.
The submaps are immediately transmitted to the mapping server where they are preprocessed and concatenated to the corresponding previously transmitted submap from the same robot.
Bookkeeping is done by duplicating the last vertex of each submap into the next submap when splitting. This also avoids discontinuities in edges and feature tracks.
In parallel the server continuously loop closes maps from different robots into a globally consistent map.
Notably, the server and console share the same code base, therefore any new functionalities can easily be integrated into either one.

\subsubsection{Submap Preprocessing}
The incoming submaps are not merged directly but rather first processed individually to ensure local accuracy.
Specifically, a configurable set of operations is executed on each robot's submap, which includes local map optimization (full bundle adjustment over all sensor data and constraints), feature quality evaluation, and intra-map loop closure (visual and LiDAR depending on what is available).
Since the submaps are processed independently of each other, the mapping server can efficiently process multiple maps concurrently.
After completion of the preprocessing steps, each finished submap is concatenated to the previous submap from the same robot.

\subsubsection{Multi-Robot Processing}
The mapping server continuously operates on the global multi-robot map and executes a second set of configurable operations (loop closure, feature quality evaluation, bundle adjustment, visualizations, absolute constraint outlier rejection, \textit{etc.}).
Here, the loop closure algorithms (visual or LiDAR) try and place all the different robots into the same reference frame and correct drift.
In contrast to the preprocessing step, the operations on the multi-robot map are always performed at a global scope, \textit{e.g.}, loop closures are detected in an intra- and inter-robot approach, and the global optimization is done jointly over all robot maps.
The collaboratively built global map can also be transmitted back to the robots to increase the accuracy of their onboard estimation~\cite{bernreiter2022collaborative}.
The increased awareness of the environment not only benefits localization accuracy but also other tasks such as global path planning.

\subsection{Offline Console}
\label{sec:console}
The offline console was ported over from \textit{maplab}, with old tools adapted to the new features regarding sensors and modalities.
There are tools for further processing maps, such as batch optimization, merging maps from different sessions, outlier rejection, key-framing, map sparsification~\cite{Dymczyk2015}, \textit{etc.}
Loop closure using LiDAR is also now possible with a new module that includes an implementation of ICP~\cite{Pomerleau2014} and G-ICP~\cite{segal2009generalized} but is not limited to these and can easily be extended.
Transformations computed by the registration module are added as loop closure edges with switchable constraints (see Section~\ref{sec:map-structure}).
For each sensor and method combination we use a predefined fixed covariance, which is set separately for each translation and rotation component.
The values are empirically chosen based on the sensor noise and the accuracy of the registration method.
Dense reconstruction can also be done using the integrated Voxblox~\cite{Oleynikova2017} plugin.
The console additionally provides tools for resource management (manipulating or visualizing the attached point clouds, images, and semantic measurements) or exporting map data (poses, IMU biases, landmarks, \textit{etc.}).

Finally, the console enables easy extensions through plugins that can run code offline and are independent of the map-building process. 
We used plugins, for example, to implement the LiDAR registration module mentioned above and a semantic loop closure module (see Section~\ref{sec:semantics}).

\begin{figure*}[!htb]
\centering
	\includegraphics[width=0.7\linewidth, trim= 0mm 0mm 0mm 0mm, clip]{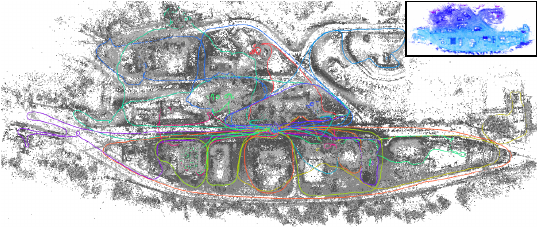}
	\caption{Visual mapping results of the global multi-robot map comprising 23 mapping runs. Individual colors denote robot trajectories and gray points denote triangulated BRISK landmarks in the multi-robot map. The top right image shows the LiDAR map by reprojecting the point clouds onto the optimized poses.}
	\label{fig:exp:arche}
	\vspace{-0.1cm}
\end{figure*}
\section{Use-Cases}
We conducted several experiments to evaluate our proposed framework thoroughly and to demonstrate its ease of use and high flexibility. 
Specifically, this section presents results on four datasets to showcase the new features and capabilities of \maplabns.
First, we validate the performance and accuracy of our proposed framework on the public HILTI SLAM 2021 dataset~\cite{hilti2021} and compare it to well-known state-of-the-art approaches.
Next, we demonstrate the real-world applicability of our proposed framework and showcase the large-scale multi-robot multi-session capabilities.
Then, we show the versatility of the landmark system by incorporating 3D LiDAR features detected from projected point clouds.
Finally, we showcase a semantic loop closure module on a custom indoor dataset.
All datasets are collected with hardware time-synchronized sensor setups.

\subsection{Validation and Comparison}
\label{sec:hilti}
We use the HILTI SLAM Challenge 2021 dataset~\cite{hilti2021} to compare our proposed framework to state-of-the-art approaches.
The dataset includes 12 recordings covering indoor office environments and challenging outdoor construction sites. 
In our experiments we exclude three sequences that are too small and do not present interesting challenges.
Three pairs (Construction Site, Basement, and Campus) of the remaining nine sequences were taken in the same environment and can be co-localized to increase accuracy.

For \maplabns, we can use the five cameras, the ADIS IMU and the OS0-64 LiDAR provided in the dataset.
We show three use-cases for \maplab using three different odometry sources: ROVIO~\cite{bloesch2017iterated}, OKVIS~\cite{leutenegger2015keyframe}, and FAST-LIO2~\cite{xu2022fast}.
Besides the standard BRISK~\cite{Leutenegger2011} descriptors, we use the external interfaces from Section~\ref{sec:landmarks} to also include SuperPoint~\cite{Detone2018SuperPoint:Description} features with SuperGlue~\cite{sarlin2020superglue} tracking, and SIFT features~\cite{lowe1999object} with LK tracking~\cite{Lucas1981}.
To reduce map size and speed up the descriptor search, we compress the SuperPoint and SIFT features using \ac{pca} from $256$ floating points to $32$.
Global loop closures are computed using all available features, and matched landmarks are merged.
We also use ICP~\cite{Pomerleau2014} from the LiDAR registration module (see Section~\ref{sec:console}) to refine our poses locally.
Covariances for the loop closure edges are predefined empirically.
Visualizations from sequence \textit{Office Mitte} are presented in Figure~\ref{fig:hilti-office-mitte}a-c.
It can be observed how  SuperPoint features better follow the building structure compared to ORB\reviewchanges{\sout{ ones}}.

\begin{table*}[!t]
    \centering
       \begin{tabular}{cccccccc|c|cc|c}
        \toprule
        \multicolumn{12}{c}{\textbf{HILTI 2021 SLAM Dataset}} \\ \midrule
        \multirow{4}{*}{\textbf{Sequence}} &  &  &  &  & &  & & \multicolumn{4}{c}{\textbf{\maplabns}}  \\
        & \textbf{ORB} & \textbf{LVI} & \textbf{RTAB} &  \multirow{2}{*}{\textbf{maplab}} & \multirow{2}{*}{\textbf{ROVIO}}& \multirow{2}{*}{\textbf{OKVIS}}& \textbf{FAST} & ROVIO & \multicolumn{2}{c|}{OKVIS} & FAST-LIO2\\
        & \textbf{SLAM3} & \textbf{SAM} & \textbf{Map} & &  & & \textbf{LIO2} & + SIFT & + SP + B & + ICP & + SP + B \\ 
        & \iconstereo\iconimu & \iconstereo\iconlidar\iconimu & \iconstereo\iconlidar\iconimu & \iconstereo\iconimu & \iconmono\iconimu & \iconstereo\iconimu & \iconlidar\iconimu & \iconstereo\iconimu & \iconstereo\iconimu & \iconstereo\iconlidar\iconimu & \iconstereo\iconlidar\iconimu \\ \midrule
        Construction 1      & 1.55\m & 0.13\m & 0.36\m & 0.16\m & 0.98\m & 1.17\m & \textbf{0.04\m} & 0.14\m & 0.08\m & 0.08\m & \textbf{0.04\m} \\
        Construction 2      & 2.77\m & 0.33\m & 0.67\m & 0.57\m & 1.50\m & 2.13\m & \textbf{0.07\m} & 0.34\m & 0.19\m & 0.19\m & \textbf{0.07\m} \\
        IC Office           & 1.86\m & 0.12\m & 1.50\m & 0.09\m & 1.16\m & 1.27\m & 0.08\m & 0.08\m & 0.08\m & \textbf{0.07\m} & \textbf{0.07\m} \\
        Office Mitte        & 1.70\m & 0.24\m & 0.94\m & 3.18\m & 0.86\m & 1.15\m & 0.12\m & 0.27\m & 0.18\m & 0.15\m & \textbf{0.10\m} \\
        Basement 3          & 1.55\m & 0.10\m & 0.38\m & 0.09\m & 3.05\m & 1.01\m & \textbf{0.05\m} & 0.09\m & 0.09\m & 0.08\m & \textbf{0.05\m} \\
        Basement 4          & 1.71\m & 0.13\m & 0.38\m & 0.11\m & 2.90\m & 1.23\m & \textbf{0.04\m} & 0.11\m & 0.10\m & 0.09\m & \textbf{0.04\m} \\
        Parking             & 5.49\m & 4.43\m & 7.82\m & 0.39\m & 6.13\m & 3.36\m & 5.00\m & 0.31\m & \textbf{0.21\m} & \textbf{0.21\m} & \textbf{0.21\m} \\
        Campus 1            & 1.93\m & 0.12\m & 0.93\m & 0.60\m & 5.10\m & 2.41\m & \textbf{0.07\m} & 0.38\m & 0.19\m & 0.17\m & \textbf{0.07\m} \\
        Campus 2            & 2.24\m & 0.14\m & 0.79\m & 0.47\m & 2.02\m & 2.23\m & 0.09\m & 0.28\m & 0.20\m & 0.18\m & \textbf{0.08\m} \\
        \midrule
        \textbf{\reviewchanges{\sout{Average}Total} Time} & 61\,\si{\min} & 68\,\si{\min} & 163\,\si{\min} & 82\,\si{\min} & 58\,\si{\min} & 121\,\si{\min} & 52\,\si{\min} & 98\,\si{\min} & 236\,\si{\min} & 267\,\si{\min} & 194\,\si{\min} \\
        \bottomrule
    \end{tabular}
    \caption{Comparison of state-of-the-art methods in terms of the RMSE of the \ac{ape}. SP + B represents SuperPoint and BRISK visual features. The icons represent the utilized sensors: monocular \protect\iconmonotext, multi-camera \protect\iconstereotext, LiDAR \protect\iconlidartext, and IMU \protect\iconimutext. \reviewchanges{The total duration of the dataset is 52 minutes.}}
    \label{tab:exp:sota_feature_comparison}
    \vspace{-0.4cm}
\end{table*}

Table~\ref{tab:exp:sota_feature_comparison} also shows the performance of the odometry sources alone, and other SLAM baselines (LVI-SAM~\cite{shan2021lvi}, ORB-SLAM3~\cite{campos2021orb}, RTAB-Map~\cite{labbe2019rtab}, and \textit{maplab}~\cite{Schneider2017}).
\textit{Maplab} and \maplab are the only methods able to use all five cameras \reviewchanges{for loop closures. For ROVIO and OKVIS we only use the frontal camera or stereo pair for odometry}.
Among all methods that use vision \maplab outperforms the baselines by a significant margin.
FAST-LIO2, which uses only LiDAR-Inertial, is the best baseline, outperforming even LVI-SAM, which is a vision-LiDAR-inertial fusion based on the same principles.
However, we show that we can also take the best performing method as odometry and further refine the result, especially improving significantly on the \textit{Parking} sequence over FAST-LIO2.
We also present a fusion of ROVIO and SIFT, demonstrating the versatility of \maplab for fast incremental improvements, independently of better deep-learned visual features.
Timings for all methods are also presented on a machine with an Intel i7-8700 and an Nvidia RTX 2080 GPU.

\subsection{Large-Scale Multi-Robot Multi-Session Mapping}
\label{sec:arche}
We demonstrate the applicability toward complex real-world scenarios by deploying our proposed framework in a large-scale training facility in Switzerland.
The environment features urban-like streets with buildings and harsh environments such as collapsed buildings with narrow spaces.
For this experiment, we recorded 23 individual runs with a handheld device with five cameras and an Ouster OS0-128 comprising more than two hours of data over approximately $10\km$ and multiple indoors-outdoors transitions.
Each run used OKVIS~\cite{leutenegger2015keyframe} for odometry.
The first five maps were used to build a global multi-robot map using the mapping server, and the remaining maps were merged using multi-session mapping in the console.
Consistency between all missions was enforced using global visual loop closures, and if available, additional absolute pose constraints from an RTK GPS.
Moreover, individual trajectories were refined by performing \reviewchanges{intra- and inter-mission} LiDAR registrations\reviewchanges{\sout{ between intra- and inter-mission poses}}.
Figure~\ref{fig:exp:arche} shows the final multi-robot map.

To quantitatively evaluate the multi-robot server we test on the public EuRoC benchmark~\cite{Burri2016}.
As the console and server use the same underlying mapping framework, excluding minor details such as operation ordering, the expected accuracy is the same.
We run all 11 sequences simultaneously in a multi-robot experiment using the mapping server, with ROVIO and BRISK.
Afterwards, we repeat the experiment \reviewchanges{by} sequentially \reviewchanges{processing each mission} using the mapping node and console \reviewchanges{and then merging them together\sout{for multi-session mapping}}.
Both scenarios achieve an average RMSE \ac{ape} of 0.043\m, but the \reviewchanges{parallelized} mapping server only takes 3\,\si{\min} 27\,\si{\s} \reviewchanges{for everything}, as opposed to 35\,\si{\min} 56\,\si{\s} for the \reviewchanges{sequential} multi-session workflow.
\reviewchanges{For both scenarios the timings include the odometry, optimization and map merging.}

\subsection{Visual Tracking on Projected LiDAR Images}
\begin{figure}[!t]
\centering
     \begin{subfigure}[b]{\columnwidth}
         \centering
         \includegraphics[width=\columnwidth]{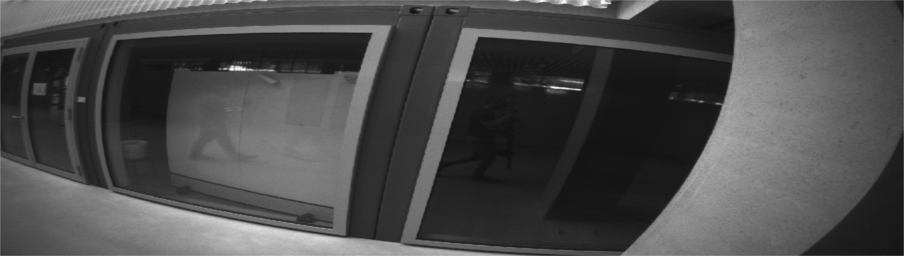}
         \caption{}
     \end{subfigure}
     \begin{subfigure}[b]{\columnwidth}
         \centering
         \includegraphics[width=\columnwidth]{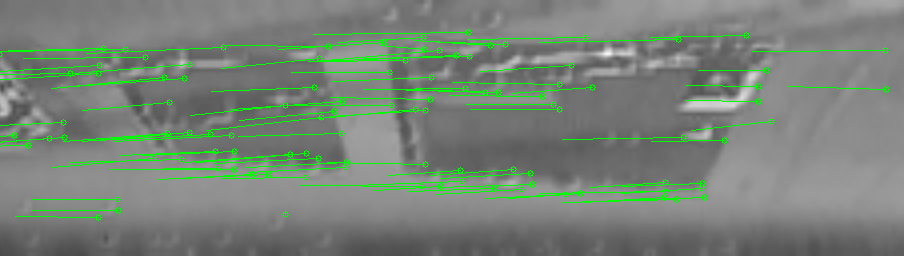}
         \caption{}
     \end{subfigure}
     \caption{Tracking keypoints on LiDAR images. (a) shows a related camera image only for illustration purposes. (b) shows the LiDAR image (cropped for visualization $\sim 40^{\circ}$) where the green circles and lines represent SuperPoint detections along with their tracked motion to the previous frame.}
     \label{fig:lidar-superpoint}
     \vspace{-0.3cm}
\end{figure}

To showcase the flexibility of the landmark system in \maplabns, we integrate 3D LiDAR keypoints\footnote{Please note that a similar workflow could be implemented for other modalities, \textit{e.g.}, RGB-D cameras, radar or sonar imaging.}.
We draw inspiration from the work of Streiff \textit{et al.}~\cite{streiff20213d3l} and project the LiDAR point cloud onto a 2D plane.
We normalize the LiDAR range and intensity values using a logarithmic scale and merge the two channels using Mertens fusion~\cite{mertens2007exposure}.
Missing pixels from bad LiDAR returns are in-painted using neighboring values.
An example image of the resulting 2D projection is shown in Figure~\ref{fig:lidar-superpoint}, alongside a camera image from the same perspective showing the environment.
We then treat the LiDAR image like a camera image and apply SuperPoint with SuperGlue to obtain point features and tracks, as shown in Figure~\ref{fig:lidar-superpoint}.
Since, for each feature observation, we have depth information from the LiDAR, we can more efficiently initialize and loop close these 3D LiDAR landmarks, as described in Section~\ref{sec:landmarks}.
Finally, we use these LiDAR keypoints to map out one of the sequences in the HILTI 2021 dataset and visualize the resulting map in Figure~\ref{fig:hilti-lidar-kps}.
The LiDAR landmarks are mapped more accurately onto the structure than the visual keypoints, as seen from the straightness of the walls.
However, they also suffer from outliers caused by noise in the LiDAR image from missing points or moving objects in the environment.

\subsection{Semantic-based Mapping}
\label{sec:semantics}
This section showcases the extensibility and modular design of \maplab by augmenting the map with semantic information and illustrating its potential application in a real-world scenario.
Initially, semantic objects are detected in an image using Mask R-CNN~\cite{He2017}, and for each detection, we use NetVLAD~\cite{Arandjelovic2018NetVLAD:Recognition} to extract a descriptor on the masked instance segmentation.
Instead of the built-in tracker, all detected objects are tracked using Deep SORT~\cite{wojke2017simple}, which extends typical spatial data association metrics with an appearance term that can directly utilize the previously extracted object descriptor.
Similar to visual landmarks, semantic objects are 3D landmarks in the \maplab map but have an associated class label and can be used for, \textit{e.g.,} semantic loop closure detection.

Finally, candidate semantic loop closures are found by directly comparing the object descriptors of the same class. First, a unique visibility filter is applied, \textit{i.e.}, two landmarks observed in a single image cannot be matched.
After geometrically verifying the candidates and clustering co-visible landmarks, a 6-DoF constraint between the two robot vertices closest to two matched landmark clusters is constructed using the relative coordinate transformation between two 3D landmark clusters\reviewchanges{, obtained through Horn's method~\cite{horn1987closed}}.
Finally, the covariance of the corresponding factor-graph constraint is calculated using the method proposed by Manoj \textit{et al.}~\cite{Manoj2015ACovariance}.

\begin{figure}[!t]
\centering
     \begin{subfigure}[b]{0.49\columnwidth}
         \centering
         \includegraphics[width=\textwidth]{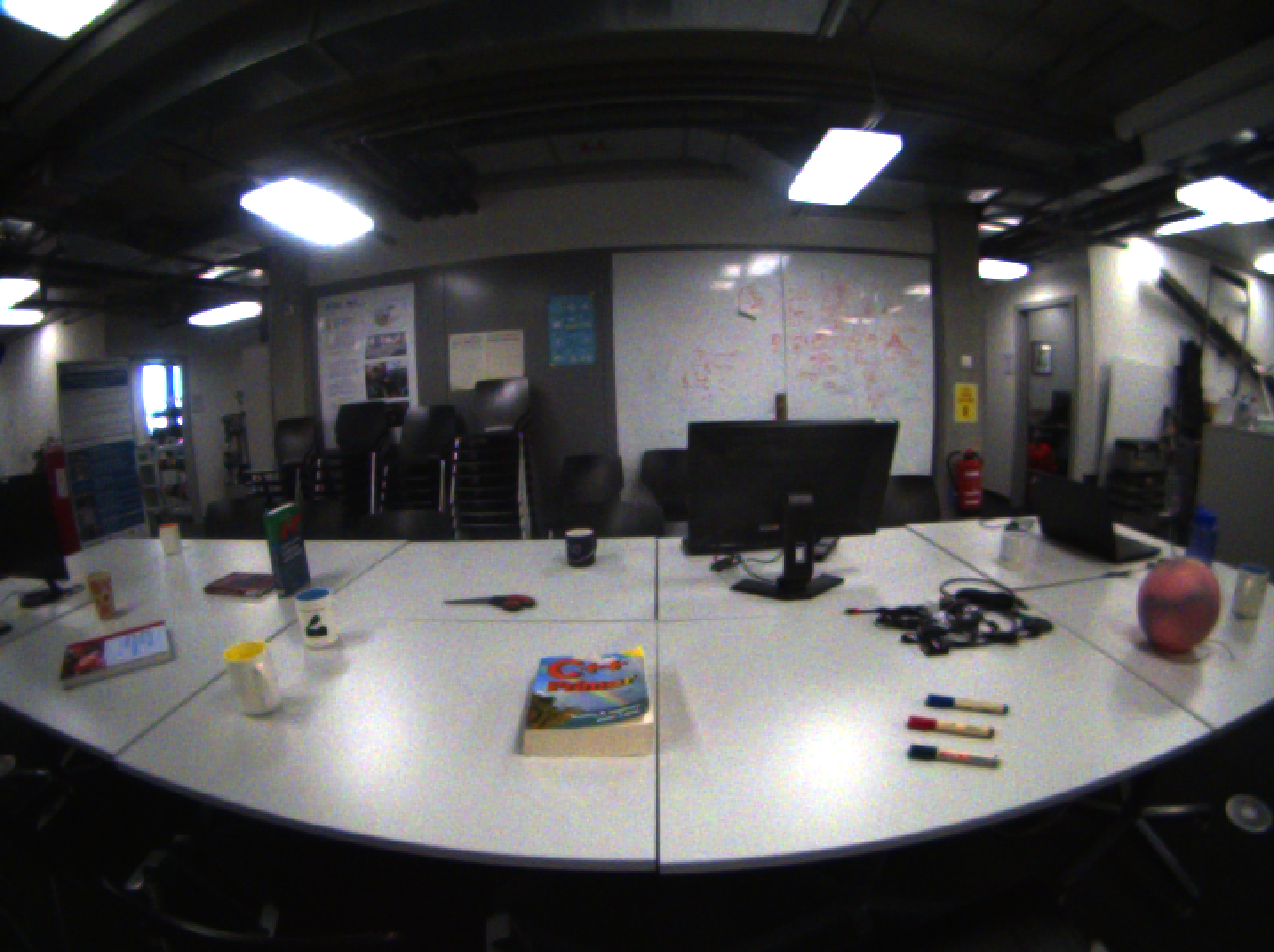}
         \caption{}
         \label{fig:semantic:A}
     \end{subfigure}
     \hfill
     \begin{subfigure}[b]{0.49\columnwidth}
         \centering
         \includegraphics[width=\textwidth]{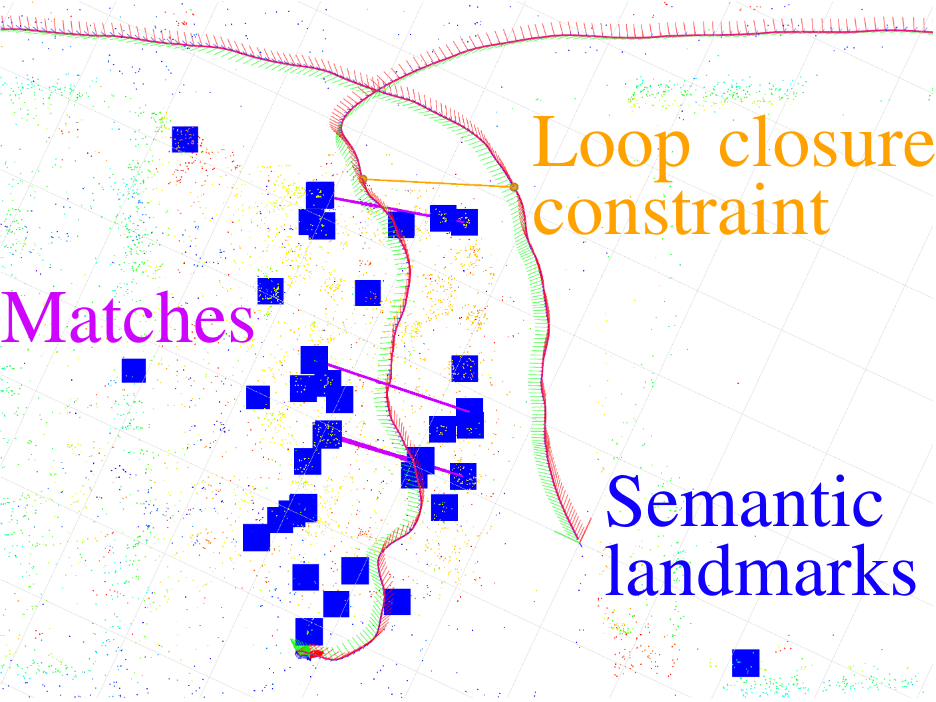}
         \caption{}
         \label{fig:semantic:B}
     \end{subfigure}
     \\
     \begin{subfigure}[b]{\columnwidth}
         \centering
         \includegraphics[width=\textwidth]{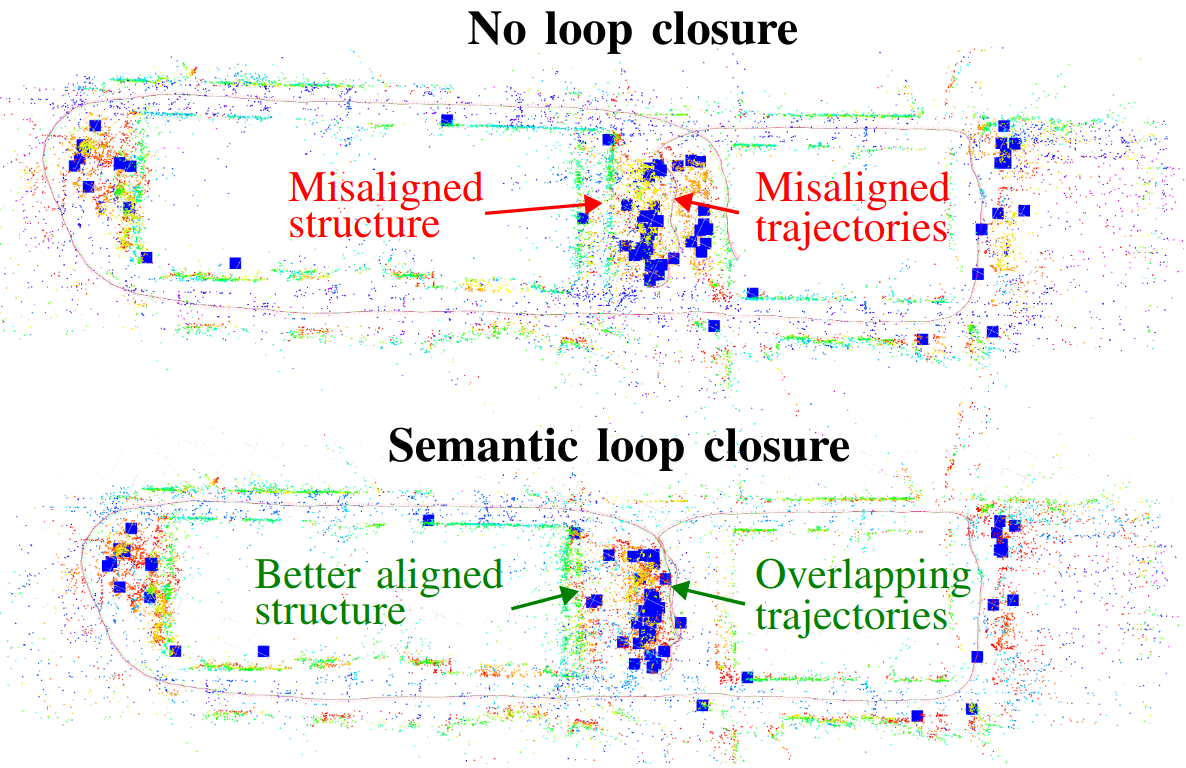}
         \caption{}
         \label{fig:semantic:C}
     \end{subfigure}
     \caption{Semantic mapping pipeline. (a) Experimental setup: a table with multiple semantic objects. (b) Loop closure matches (magenta) between semantic landmarks (blue), resulting in a loop closure constraint (orange). (c) Visual-inertial-semantic map before and after semantic loop closure.}
     \vspace{-0.3cm}
\end{figure}

We collected an indoor dataset in an office environment with multiple objects using an RGB-inertial sensor~\cite{Tschopp2020}. 
We observe an office table with objects on two occasions while leaving some time to accumulate drift (see  Figure~\ref{fig:semantic:A} and~\ref{fig:semantic:C}). 
Figure~\ref{fig:semantic:B} shows semantic landmark clusters and detected loop closure candidates. 
After adding the loop closure edges from the semantic objects to the full factor graph, the drift significantly reduces, and an improved map can be seen in Figure~\ref{fig:semantic:C}.

\section{Conclusion}
We presented a research platform for multi-modal and multi-robot mapping, supporting online and offline processing of the maps.
We showcase state-of-the-art performance on a large-scale SLAM benchmark and multiple experimental use cases for \maplabns.
Our proposed mapping framework's flexible and modular design facilitates research in various robotic applications and yields important implications in academia and industry.
%
%
The code and tutorials to reproduce the experiments \reviewchanges{\sout{will be}are} available on the wiki of the repository.

\begin{acronym}
\acro{flann}[FLANN]{Fast Library for Approximate Nearest Neighbors}
\acro{adas}[ADAS]{advanced driving assistance systems}
\acro{ae}[AE]{auto-exposure}
\acro{asl}[ASL]{Autonomous Systems Lab}
\acro{ba}[BA]{bundle adjustment}
\acro{bm}[BM]{block-matching}
\acro{bow}[BoW]{Bag-of-Words}
\acro{brisk}[BRISK]{Binary Rotation Invariant Scalable Keypoint}
\acro{clahe}[CLAHE]{contrast limiting adaptive histogram equalization}
\acro{cnn}[CNN]{Convolutional Neural Network}
\acro{cpu}[CPU]{central processing unit}
\acro{davis}[DAVIS]{Dynamic and Active Vision Sensor}
\acro{dcnn}[DCNN]{Deep Convolutional Neural Network}
\acro{dl}[DL]{deep learning}
\acro{dof}[DoF]{degrees of freedom}
\acro{dso}[DSO]{Direct Sparse Odometry}
\acro{dvs}[DVS]{Dynamic Vision Sensor}
\acro{ekf}[EKF]{extended Kalman filter}
\acro{etcs}[ETCS]{European Train Control System}
\acro{etsc}[ETSC]{European Train Security Council}
\acro{esdf}[ESDF]{Euclidean signed distance field}
\acro{fast}[FAST]{Features form Accelerated Segment Test}
\acro{fc}[FC]{fully connected}
\acro{fir}[FIR]{finite impulse response}
\acro{fov}[FoV]{field of view}
\acro{fpga}[FPGA]{field-programmable gate array}
\acro{fps}[FPS]{frames per second}
\acro{gnss}[GNSS]{global navigation satellite system}
\acro{gp}[GP]{Gaussian Process}
\acro{gpm}[GPM]{Gaussian preintegrated measurement}
\acro{gps}[GPS]{Global Positioning System}
\acro{gpu}[GPU]{graphics processing unit}
\acro{gt}[GT]{ground truth}
\acro{gtc}[GTC]{ground truth clustering}
\acro{gtsam}[GTSAM]{Georgia Tech Smoothing and Mapping library}
\acro{hdr}[HDR]{High Dynamic Range}
\acro{hs}[HS]{Hough space}
\acro{ht}[HT]{Hough transform}
\acro{i2c}[I$^2$C]{Inter-Integrated Circuit}
\acro{idol}[IDOL]{IMU-DVS  Odometry  with Lines}
\acro{imu}[IMU]{inertial measurement unit}
\acro{ins}[INS]{inertial navigation system}
\acro{iou}[IOU]{intersection over union}
\acro{kf}[KF]{Kalman filter}
\acro{led}[LED]{light emitting diode}
\acro{lidar}[LiDAR]{Light Detection and Ranging sensor}
\acro{lsd}[LSD]{line segment detector}
\acro{lssvm}[LSSVM]{least squares support vector machine}
\acro{lwir}[LWIR]{long-wave infrared}
\acro{mav}[MAV]{micro aerial vehicle}
\acro{mcu}[MCU]{micro controller unit}
\acro{nclt}[NCLT]{North Campus Long-Term}
\acro{nmi}[NMI]{normalized mutual information}
\acro{nms}[NMS]{non-maxima suppression}
\acro{nn}[NN]{nearest neighbor}
\acro{os}[OS]{operating system}
\acro{pcb}[PCB]{printed circuit board}
\acro{pc}[PC]{point cloud}
\acro{pca}[PCA]{principal component analysis}
\acro{pcm}[PCM]{probabilistic curvemap}
\acro{pf}[PF]{particle filter}
\acro{pps}[PPS]{Pulse per second}
\acro{ptp}[PTP]{Precision Time Protocol}
\acro{ransac}[RANSAC]{random sample consensus}
\acro{rgbdi}[RGB-D-I]{Color-Depth-Inertial}
\acro{riou}[R-IOU]{reprojection intersection over union}
\acro{rmse}[RMSE]{root mean square error}
\acro{rnn}[RNN]{recurrent neural network}
\acro{roi}[ROI]{region of interest}
\acro{ros}[ROS]{Robot Operating System}
\acro{rqe}[RQE]{Rényi's Quadric Entropy}
\acro{rtk}[RTK]{real time kinematics}
\acro{sbb}[SBB]{Schweizerische Bundesbahnen}
\acro{sift}[SIFT]{Scale Invariant Feature Transform}
\acro{slam}[SLAM]{Simultaneous Localization And Mapping}
\acro{snn}[SNN]{spiking neural network}
\acro{snr}[SNR]{signal to noise ratio}
\acro{sp}[SP]{SuperPoint}
\acro{spi}[SPI]{Serial Peripheral Interface}
\acro{sq}[SQ]{superquadric}
\acro{swe}[SWE]{Sliding Window Estimator}
\acro{tof}[ToF]{time of flight}
\acro{tsdf}[TSDF]{Truncated Signed Distance Function}
\acro{tum}[TUM]{Technische Universit\"at M\"unchen}
\acro{uart}[UART]{Universal Asynchronous Receiver Transmitter}
\acro{uav}[UAV]{unmanned aerial vehicle}
\acro{usb}[USB]{universal serial bus}
\acro{vbg}[VBG]{Verkehrsbetriebe Glattal AG}
\acro{vbz}[VBZ]{Verkehrsbetriebe Zürich}
\acro{VersaVIS}[VersaVIS]{Open Versatile Multi-Camera Visual-Inertial Sensor Suite}
\acro{vi}[VI]{visual-inertial}
\acro{vio}[VIO]{visual-inertial odometry}
\acro{vo}[VO]{visual odometry}
\acro{zvv}[ZVV]{Z\"urich Verkehrsverein}
\acro{svd}[SVD]{Singular Value Decomposition}
\acro{dlt}[DLT]{Direct Linear Transform}
\acro{ugv}[UGV]{unmanned ground vehicle}
\acro{ar}[AR]{augmented reality}
\acro{vr}[VR]{virtual reality}
\acro{ato}[ATO]{autonomous train operation}
\acro{ape}[APE]{absolute position error}
\end{acronym}

\bibliographystyle{IEEEtran}
\bibliography{maplab.bib}

\end{document}